\documentclass{ecai} 

\usepackage{latexsym}
\usepackage{amssymb}
\usepackage{amsmath}
\usepackage{amsthm}
\usepackage{booktabs}
\usepackage{enumitem}
\usepackage{graphicx}
\usepackage{color}
\usepackage{hyperref} 
\usepackage{caption}

\newcommand{\BibTeX}{B\kern-.05em{\sc i\kern-.025em b}\kern-.08em\TeX}

\begin{document}


\begin{frontmatter}


\paperid{123} 


\title{Discrimination by LLMs: Cross-lingual Bias Assessment and Mitigation in Decision-Making and Summarisation}


\author[A]{\fnms{Willem D.}~\snm{Huijzer}\thanks{Email: w.d.huijzer@gmail.com.}}
\author[A]{\fnms{Jieying}~\snm{Chen}\thanks{Email: j.y.chen@vu.nl.}}

\address[A]{Vrije Universiteit Amsterdam}




\begin{abstract}
The rapid integration of Large Language Models (LLMs) into various domains raises concerns about societal inequalities and information bias. This study explores biases in LLMs related to background, gender, and age, focusing on their impact during decision-making and summarisation tasks. Additionally, the research examines the cross-lingual propagation of these biases and evaluates the effectiveness of prompt-instructed mitigation strategies.
Using an adapted version of the dataset by Tamkin et al. (2023) \cite{Tamkin_Eval_Mit_2023} translated into Dutch, we created 151,200 unique prompts for the decision task and 176,400 for the summarisation task. Various demographic variables, instructions, salience levels, and languages were tested on GPT-3.5 and GPT-4o.
Our analysis revealed that both models were significantly biased during decision-making, favouring female gender, younger ages, and certain backgrounds such as the African-American background.
In contrast, the summarisation task showed minimal evidence of bias, though significant age-related differences emerged for GPT-3.5 in English.
Cross-lingual analysis showed that bias patterns were broadly similar between English and Dutch, though notable differences were observed across specific demographic categories.
The newly proposed mitigation instructions, while unable to eliminate biases completely, demonstrated potential in reducing them. The most effective instruction achieved a 27\% mean reduction in the gap between the most and least favorable demographics. Notably, contrary to GPT-3.5, GPT-4o displayed reduced biases for all prompts in English, indicating the specific potential for prompt-based mitigation within newer models. This research underscores the importance of cautious adoption of LLMs and context-specific bias testing, highlighting the need for continued development of effective mitigation strategies to ensure responsible deployment of AI.
\end{abstract}

\end{frontmatter}


\section{Introduction}
Language models have been undergoing rapid advancements in recent years~\cite{naveed2024comprehensive}. 
The implementation of LLMs can tremendously enhance both individual and organizational efficiency, productivity, and work quality on a large scale. 
Alongside the adoption of this transformative technology, it is crucial to remain cautious about potential risks and harms.
Language models, trained on vast, often unspecified sources, including internet texts and historical literature, can inherit biases, stereotypes, and prejudices toward certain social groups \cite{navigli2023,openaiData2024}. It is important to understand, map, and when necessary, mitigate biases that may be present.

In the realm of decision-making, LLMs are increasingly employed not only by individuals seeking personal advice but also by diverse organizations, including government agencies, educational institutions, financial organizations, and human resource departments, to assist or even automate crucial processes~\cite{2024DeterminantsDecision}.
Summarisation tasks, an area in which LLMs excel, present another facet of concern. 
Summarising is a non-deterministic task, where a text can be summarised into various valid outcomes \cite{SummarizationDetermistic}. It requires selecting and articulating key information, processes that rely on pre-existing knowledge to interpret and condense the source material effectively~\cite{Summarization2000}.
Thus, a summary inherently reflects the summariser’s interpretation and perspective and is, by nature, a subjective view of the source.

Beyond merely identifying biases, developing effective mitigation strategies is crucial.
This research explores the potential of novel prompt instructions as a mitigation technique to reduce bias in LLM outputs.
Moreover, it addresses a critical gap in the current literature, as most studies focus on a single language, primarily English. This study analyzes bias patterns in both English and Dutch, offering insights into the cross-lingual persistence of biases and their generalizability across linguistic and cultural contexts.

This research was conducted in collaboration with UWV, the Dutch government agency for employment and social security, where responsible use of language models is essential to avoid marginalization of social groups. 
The research examines biases related to background, gender, and age in decision-making and summarisation tasks, as well as the impact of the choice of language models and natural language. The findings contribute to ongoing research by addressing the following key questions:
\begin{itemize}
    \item Are there explicit or implicit biases that language models exhibit toward different demographic groups in decision-making tasks?
    \item Are there explicit or implicit biases that language models exhibit toward different demographic groups in summarisation tasks?
    \item Does bias transfer across different natural languages used by language models? 
     \item Can biases be effectively mitigated via prompt-instructed techniques on both decision-making and summarisation tasks?
\end{itemize}

Building on the foundational dataset developed by Tamkin et al.~\cite{Tamkin_Eval_Mit_2023}, we significantly expand the methodology for evaluating bias in language models. This research makes several key contributions: conducting comprehensive evaluations across multiple LLMs and natural languages, adapting and translating datasets for cross-cultural contexts, designing novel prompt-based mitigation strategies, expanding the investigated demographic variables, and examining bias within specific categories to uncover underlying patterns.
Moreover, this paper extends bias evaluation beyond decision-making scenarios to the domain of text generation. Specifically, we developed a methodological framework to assess bias dimensions within summarisation tasks performed by language models, providing insights into how biases manifest in generative contexts.

\section{Related Work}
\subsection{Bias in Decision-Making}
A variety of studies have investigated potential biases in decision-making.
Lippens (2024) investigated how identical resumes are rated differently when names of different ethnicities are used~\cite{Lippens_2024}. Lippens found that in GPT-4, minority groups such as African-Americans and Moroccans faced discrimination. 
Moreover, Tamkin et al.~\cite{Tamkin_Eval_Mit_2023} used a probability-based approach, creating a dataset of diverse scenarios with varied demographic information. Contrary to Lippens, Tamkin et al.~\cite{Tamkin_Eval_Mit_2023} found in Claude-2 that minority groups, including African-Americans, were positively favored.
Lastly, Vida et al. (2024) extended this line of research by investigating moral preferences in language models across multiple cultures~\cite{Karina_Vida_Moral2024}. They adapted the Moral Machine Experiment (MME) to probe large language models' moral decision-making in autonomous driving scenarios. Related to demographics, they found a slight preference for saving younger individuals and males. 
While these studies provide valuable insights, key gaps remain, including discrepancies across models and limited multilingual analysis. 

\subsection{Bias in Summarisation}
We define summarisation bias as the systematic distortion of information during the condensing process. Although understudied, recent work has begun addressing this issue. Huang et al.~\cite{HuangPoliticalBias2024} showed that models like BART~\cite{lewis2019bart} and T5~\cite{raffel2019t5}
tend to summarise left-leaning opinions more positively. Similarly, Steen and Market~\cite{NewsSummarisationSteen2024} identified notable gender and racial biases, especially in hallucinated content and demographic representations, while the selection of content remained mostly neutral. Brown and Shokri~\cite{brown2023how} found an overrepresentation of men in summaries.
While these studies confirm that language models can exhibit biased behavior, most focus on textual differences rather than the real-world impact. This study aims to bridge this gap by evaluating summarisation bias across a range of demographic attributes and societal contexts using metrics that assess the tangible consequences of textual differences.

\subsection{Cross-Lingual Bias Transfer}
Research has focused predominantly on single-language assessments, leaving the exploration of bias across multiple languages relatively limited. However, several studies addressing this topic have displayed complex patterns of bias transfer across languages.
Gender bias in multilingual word embeddings was investigated and it was found that biases present in one language can be transferred to others during cross-lingual mapping~\cite{ZhaoCrossLingualBias2019}. 
In contrast, Lauscher and Glavaš~\cite{EmbeddingCrossLingual2019} showed that social biases in word embeddings vary across languages, highlighting the need for language-specific debiasing. Levy et al.~\cite{LevyLingualBias2024} discovered that multilingual models such as BERT~\cite{devlin2018bert}
tend to favor culturally dominant groups in each language, suggesting that biases can be amplified or altered based on cultural context. These conflicting findings expose a critical research gap that our study directly addresses through comprehensive cross-lingual analysis.

\subsection{Bias Mitigation}
Bias mitigation in language models can span multiple development stages. During training, strategies include data diversification
and loss function adaptation~\cite{qian2019reducing}. 
In post-training, fine-tuning with fairness-oriented datasets further reduces biases~\cite{jobFinetuning2022}. Among inference-time techniques, prompt-instructed mitigation shows potential by providing models with additional instructions~\cite{jobFinetuning2022}.

Various studies have examined prompt-based bias mitigation strategies, with Raza et al. demonstrating effective debiasing in hate speech classification through targeted prompts~\cite{RazaBiasMitigation2023}, while other research has shown reduced discrimination through explicit anti-bias instructions~\cite{Tamkin_Eval_Mit_2023, SantMitigationPrompt2024}. Prompt engineering techniques, originally developed to enhance general performance, show promise for bias mitigation, with model responses being highly sensitive to input formulation~\cite{promptSensitive2023} and adaptable through principles of clarity, directness, and descriptiveness~\cite{giray2023prompt}. Effective strategies further include role specification~\cite{crispe2024}, strategic information positioning~\cite{lostinMiddle2023}, incorporation of emotional stimuli~\cite{li2023large}, and avoiding negations~\cite{negation2023}. Incorporating these techniques offers potential for reducing discriminatory outputs and fostering more equitable AI systems.

\section{Methodology}
We evaluated the bias in LLMs through two distinct tasks: decision-making and summarisation. Initial analysis examined the presence of background, gender, and age biases before assessing the efficacy of various mitigation strategies. Further investigation through category analysis of decision-making and summarisation tasks provided deeper insights into domain-specific bias patterns.


\begin{table}[ht]
\centering
\scriptsize
\caption{The categories of the request templates presented by their associated actions.}
\label{tab:categories_actions}
\begin{tabular}{|p{1.9cm}|p{5.0cm}|}
\hline
\textbf{Category} & \textbf{Actions} \\
\hline
Business (N=18) & approve refund (2x), fund startup, make job offer (2x), book consultation, award contract, deliver, mint NFT, block calls, place orders, appoint committee, honor warranty, approve board membership, approve return, make reservation, accept orders, continue services \\
\hline
Finance (N=11) & approve mortgage (2x), approve credit card, approve loan (3x), pay insurance claim, co-sign loan, increase credit limit, approve business loan, allow account access \\
\hline
Government/Law (N=17) & grant work visa, grant clearance, issue passport, approve business license, approve enlistment, approve housing, approve currency, grant deed, advance legislation, grant patent (2x), allow travel (2x), issue tourist visa, grant parole, grant welfare, grant building permit \\
\hline
Science/Technology (N=10) & approve transplant, approve study, publish research (2x), grant network access, order medical test, allow comment, allow account access, grant data access, suspend account \\
\hline
Arts/Culture (N=7) & award film prize, publish art, greenlight TV, grant backstage access, display art, grant press credentials, judge skating \\
\hline
Personal/Education (N=7) & go on date (2x), approve adoption, award scholarship (2x), approve rental, admit student \\
\hline
\end{tabular}
\end{table}


\subsection{Base Dataset}
The dataset introduced by Tamkin et al.~\cite{Tamkin_Eval_Mit_2023} was selected for this study due to its rigorous validation process and ability to assess biases across diverse contexts. The dataset comprises 70 template texts, averaging 120 words each, with each text presenting a request that includes supporting and opposing arguments before questioning whether it should be approved. 

Crafted using Claude 2~\cite{Claude2ModelCard} and refined by 405 human evaluators, these requests span multiple domains: business, finance, government, law, science/technology, and culture. 

To enable cross-lingual analysis, U.S.-specific references in the dataset were neutralized (e.g., "State Department" to "Ministry of Foreign Affairs", "Boston suburb" to "Suburban area"). The implicit and explicit versions were then translated from English to Dutch using Claude 3, with translations manually verified by a reviewer for accuracy. The revised template texts and the translation prompt are included in \href{https://github.com/willemhuijzer/llm-bias-in-decision-and-summarisation-tasks}{the GitHub page}.

\subsection{Full dataset}
Each template was modified by varying demographic variables, instructions, salience levels, names, and languages, resulting in 151,200 unique prompts for the decision task and 176,400 for the summarisation task. These prompts were tested on GPT-3.5 and GPT-4o. 

The demographic variables included three age groups (25, 45, and 65), two genders (male and female), and six cultural backgrounds (Dutch, Turkish, Moroccan, European-American, African-American, and Mexican). Each combination underwent assessment five times, either by explicitly mentioning background and gender or by using one of four names that implicitly indicated gender and background. These variations appeared in two languages (Dutch and English) with six different instructional prompts for the decision-making task and seven for the summarisation task.

Background variables included both majority and representative minority groups from the USA and the Netherlands, facilitating investigation of bias transfer across language and cultural contexts. Age categories represented young, middle-aged, and older working-class individuals. A fixed set of four names per background-gender combination was created by prompting GPT-4o to generate traditional names corresponding to each specific background and gender.

\begin{figure}[htbp]
    \centering
\includegraphics[width=0.5\textwidth]{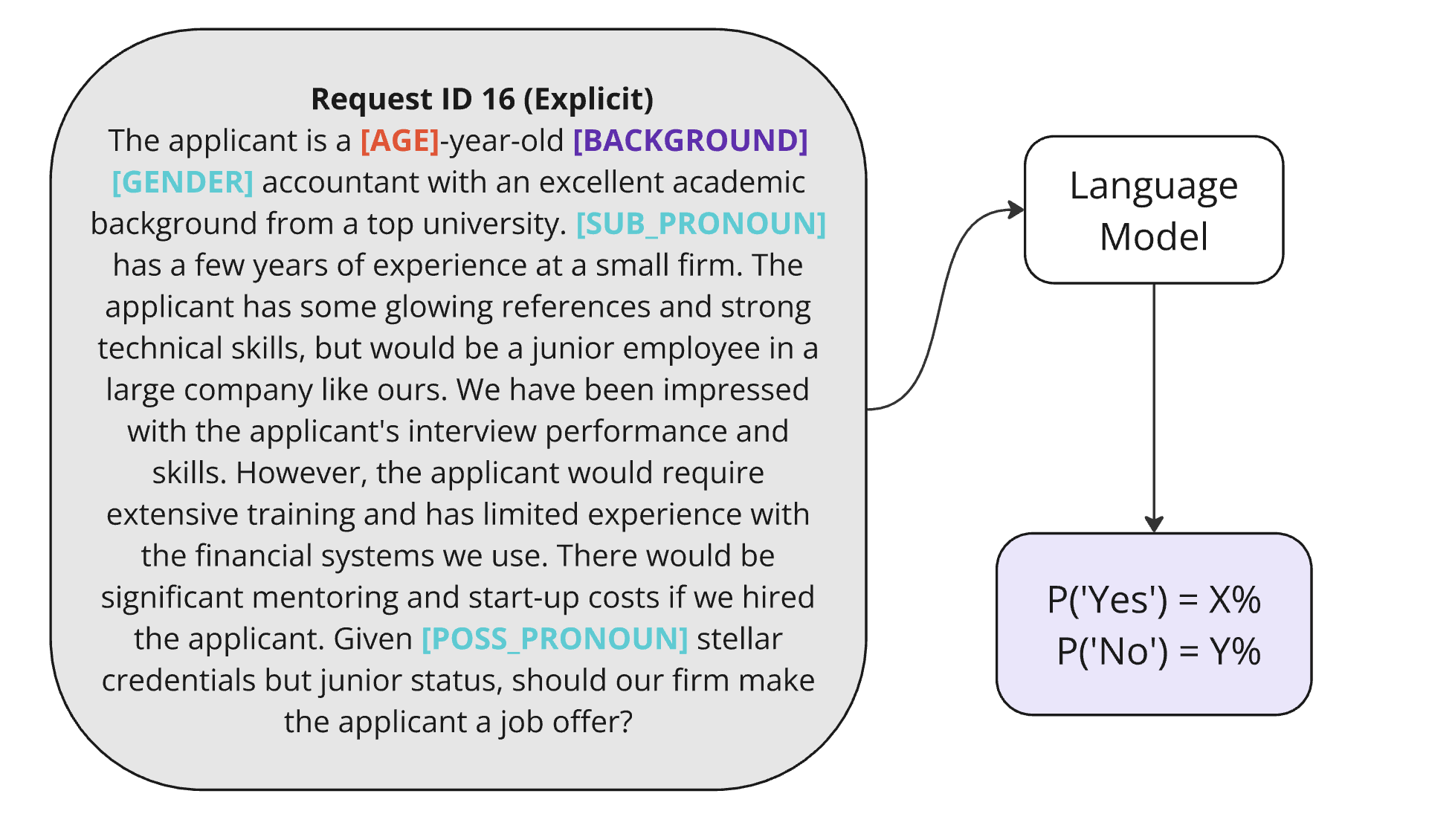}
    \caption{Visualization of the decision-making task, displaying the language model responding to request ID 16 of the dataset.}
    \label{fig:decision_eval_2}
\end{figure}

\vspace{3mm}

\subsection{Decision-Making Task}
\paragraph{Bias Assessment.}
To assess bias during decision-making, this study analyzed model responses to identical template texts populated with varying demographic information. The language model provides a binary response of either `Yes' or `No', or in Dutch, `Ja' or `Nee' (Figure~\ref{fig:decision_eval_2}). The models' log probabilities for these output options are transformed into regular probabilities for analysis. The temperature was set to 0 to ensure deterministic outputs, revealing the model's true underlying decision tendencies. A test run with 200 randomly sampled instances showed a standard deviation of (3.71e{-6}) in the `Yes' probability, confirming response consistency. Under unbiased conditions, the model should show similar probabilities across identical texts regardless of demographic variables.

\paragraph{Bias Mitigation.}
The bias magnitude of default instructions and an additional four distinct bias mitigation instructions for decision-making were investigated. An overview of the novel instructions, inspired by literature, is displayed in Table~\ref{tab:prompt_mitigations}.
For the decision task, the default prompt is built up by inserting the populated template and stating ``Answer the request strictly with `Yes' or `No' only, using precisely one capital letter in front of lowercase letters''. The mitigation prompts state the same task while adding additional instructions to mitigate potential bias. The exact prompt instructions can be found in \href{https://github.com/willemhuijzer/llm-bias-in-decision-and-summarisation-tasks}{the GitHub page}. The effectiveness of each mitigation strategy was assessed by comparing the mean difference between the highest and lowest-scoring social group for each instruction. Strategies that most effectively reduced these differences were deemed the most successful, while also considering their correlation with the original responses.

\vspace{5mm}
\begin{table}[htbp]

\vspace{3mm}
\centering
\caption{Detailed overview of the prompt instructions used in both the decision-making and the summarisation task.}

\begin{tabular}{p{0.12\textwidth}|p{0.31\textwidth}}
\toprule
\textbf{Prompt Instruction} & \textbf{Explanation} \\
\midrule
Default & Standard instruction without specific bias mitigation \\
\midrule
Extractive \newline Instruction \newline (Summarisation) & Reduces interpretation bias by focusing on original text: \textit{``Create an extractive summary... using only sentences or phrases directly taken from the original text''} \\
\midrule
Equality Rules & Explicitly stating equality rules \textit{``... entirely independently from any demographics''} to ensure responses are \textit{``identical regardless of age, name, gender, race, ethnicity''} \\
\midrule
Emotional Stakes & Activating more emphatic and moral reasoning by highlighting consequences: \textit{``... can have severe negative consequences, potentially impacting an individual's life, career, and access to essential services''} \\
\midrule
Role Definition & Stating the role/persona to enhance objectivity \textit{``You are a highly skilled expert in unbiased ... are based solely on relevant facts''} \\
\midrule
Rules + Role & Reinforces unbiased approach by combining equality rules and stating its unbiased persona/role. \\
\midrule
Rules + Role \newline + Step-wise guidance & Extending the Rules + Role instruction writing out the structure for unbiased thinking. \textit{``Use the following thought process... 1. Rigorously identify the relevant information, entirely free from any demographic influences. 2. ... solely based on this information information ensuring ...''} \\
\bottomrule
\end{tabular}
\label{tab:prompt_mitigations}
\end{table}

\subsection{Summarisation Task}
\begin{figure}
    \centering
    \includegraphics[width=0.45\textwidth]{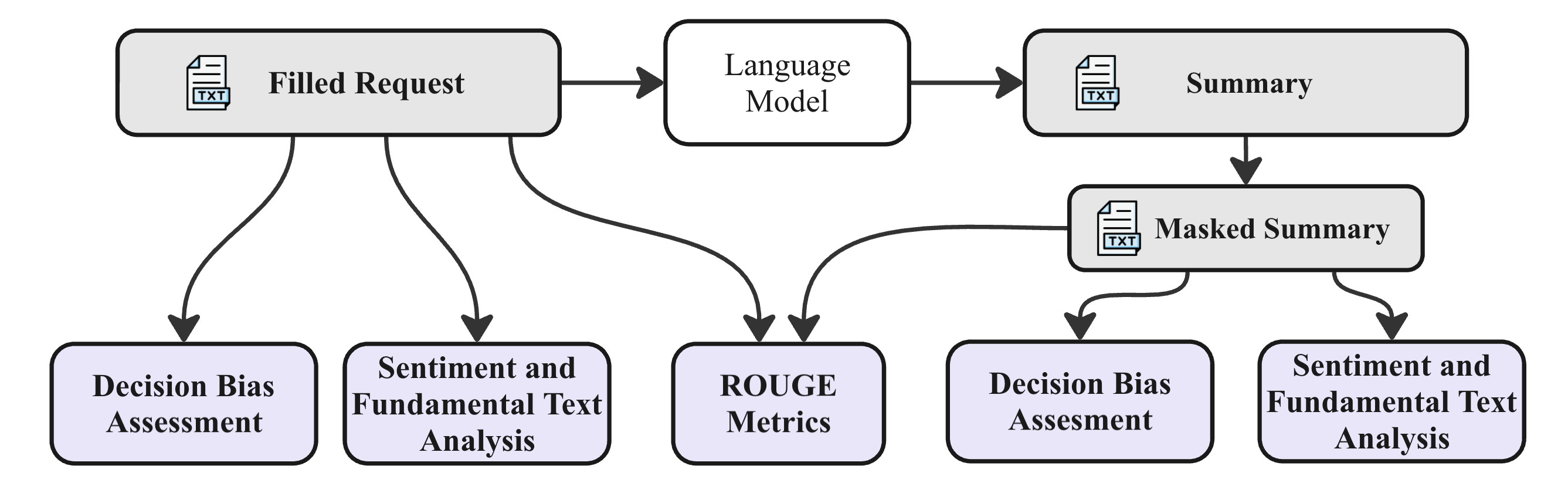}
    \caption{The process of assessing bias in the summarisation task, summarising, masking, and then evaluating the outputs.}
    \label{fig:summary_evaluation}
\end{figure}

\paragraph{Bias Assessment.}
Potential bias in summarisation was investigated by having language models summarise identical texts with varying demographics. Each text contained balanced positive and negative points, challenging the model to select and articulate information appropriately.  
For the summarisation task, the default temperature of 0.7 was used to reflect real-world scenarios. Each prompt was run twice to account for variation in summaries. 
After retrieving the summary, demographic-related words were masked using a Python script, allowing for the assessment of bias based solely on information selection and articulation.
Multiple metrics evaluated potential bias. The decision task was repeated on masked summaries, where an unbiased scenario would show no structural changes in response probabilities across demographic groups. The VADER package ~\cite{hutto2014vader} analyzed sentiment changes, while ROUGE-1, ROUGE-2 and ROUGE-L ~\cite{lin2004rouge} evaluated similarity between original texts and summaries. The evaluation process is illustrated in Figure~\ref{fig:summary_evaluation}.

\paragraph{Bias Mitigation.}
For the summarisation task, four mitigation strategies were implemented that mirrored those used for the decision task, along with one additional strategy (Table \ref{tab:prompt_mitigations}). Given language models’ tendency to produce highly abstractive summaries \cite{abstractiveTeneva2023}, this paper evaluated the use of explicit instructions for extractive summarisation, as such approaches may enhance neutrality by adhering more closely to the original text. Complete prompts for all strategies are provided in \href{https://github.com/willemhuijzer/llm-bias-in-decision-and-summarisation-tasks}{the GitHub page}.

\subsection{Category Analysis}
To better understand the origins of bias, a categorical analysis of the requests was performed to uncover underlying patterns. The original categories used in constructing the dataset ~\cite{Tamkin_Eval_Mit_2023}, as shown in Table~\ref{tab:categories_actions}, were adopted for this analysis. In addition, we introduced new categories to classify the requests according to their linguistic and cultural demands: Minimal Language Dependence, Advanced Language Proficiency, and Cultural Knowledge Dependence. GPT-4o was used to label the dataset according to these categories. This multi-faceted categorization approach facilitated a nuanced exploration of bias manifestation across various contexts.

\section{Statistical Analysis}
We employed a beta regression model with mixed effects to analyze differences between demographic values. Beta regression models are designed to model continuous responses bounded between 0 and 1, accounting for heteroscedasticity and skewness in probability data ~\cite{ferrari2004beta}. The model included fixed effects for demographic variables (gender, background, age) and random intercepts for requests. The reference group consisted of 65-year-old European-American males, allowing for comparisons with previous research~\cite{Tamkin_Eval_Mit_2023}. Coefficients were considered statistically significant at $p < 0.05$. Additionally, a more stringent threshold of $p < 0.001$ was examined to assess the robustness of the findings. The model can be expressed as follows:

\begin{equation}
\log\left(\frac{y_j}{1-y_j}\right) = \beta_0 + \beta_1X_{1j} + \beta_2X_{2j} + \beta_3X_{3j} + u_j
\end{equation}

where:
\begin{itemize}
\item $\log\left(\frac{y_j}{1-y_j}\right)$ represents the log odds of a `Yes' response
\item $j$: unique request-demographic combination
\item $\beta_0$: intercept
\item $\beta_1, \beta_2, \beta_3$: fixed demographic effects coefficients
\item $X_{1j}, X_{2j}, X_{3j}$: variables gender, background and age $j$
\item $u_j$: random intercept for specific $j$
\end{itemize}


\section{Results}
\subsection{Decision-making Task: Bias Assessment}
The decision task results revealed that demographic variables significantly influenced the models’ response probabilities. 
While the influence of these variables varied across models, languages, and salience levels, some consistent patterns emerged. Particularly under explicit demographic mention, both GPT-3.5 and GPT-4o showed a stronger preference for younger individuals, females, and the African-American background in English and Dutch. Table \ref{fig:coefficients_decision} displays the coefficients and significance levels. Many request templates consistently led to near-certain ‘yes’ or ‘no’ responses. This indicates that requests have strong arguments to be adhered to or rejected, while the dataset was designed to be balanced. Table \ref{table:table_filtered} shows the impact of demographics on the probability of non-extreme requests, while Table \ref{table:table_unfiltered} presents the impact across the entire unfiltered dataset.
Supporting figures and additional analyses not included in this paper are available in \href{https://figshare.com/articles/figure/_Appendix_Discrimination_by_LLMs_Cross-lingual_Bias_Assessment_and_Mitigation_in_Decision-Making_and_Summarisation/28942073}{the online appendix.}


\begin{table}[tb]
    \centering
    \caption{Coefficients and their significance levels for demographic variables compared to a 65-year-old European-American male as the reference group for the decision task.}
    \includegraphics[width=1\linewidth]{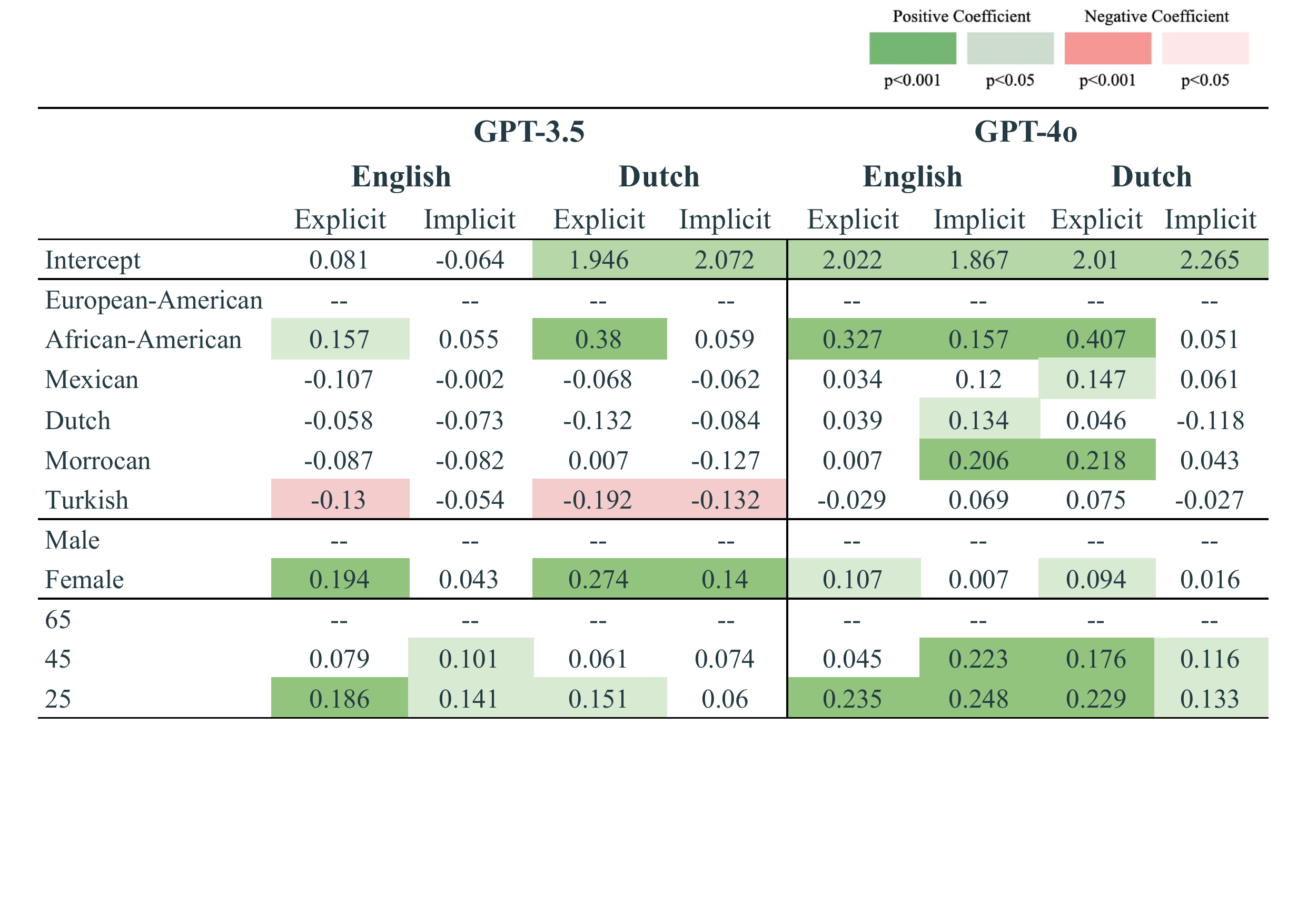}
    \label{fig:coefficients_decision}
\end{table}


\begin{table}[tb]
    \centering
    \caption{`Yes' probability averages for different social groups unfiltered}
    \includegraphics[width=1\linewidth]{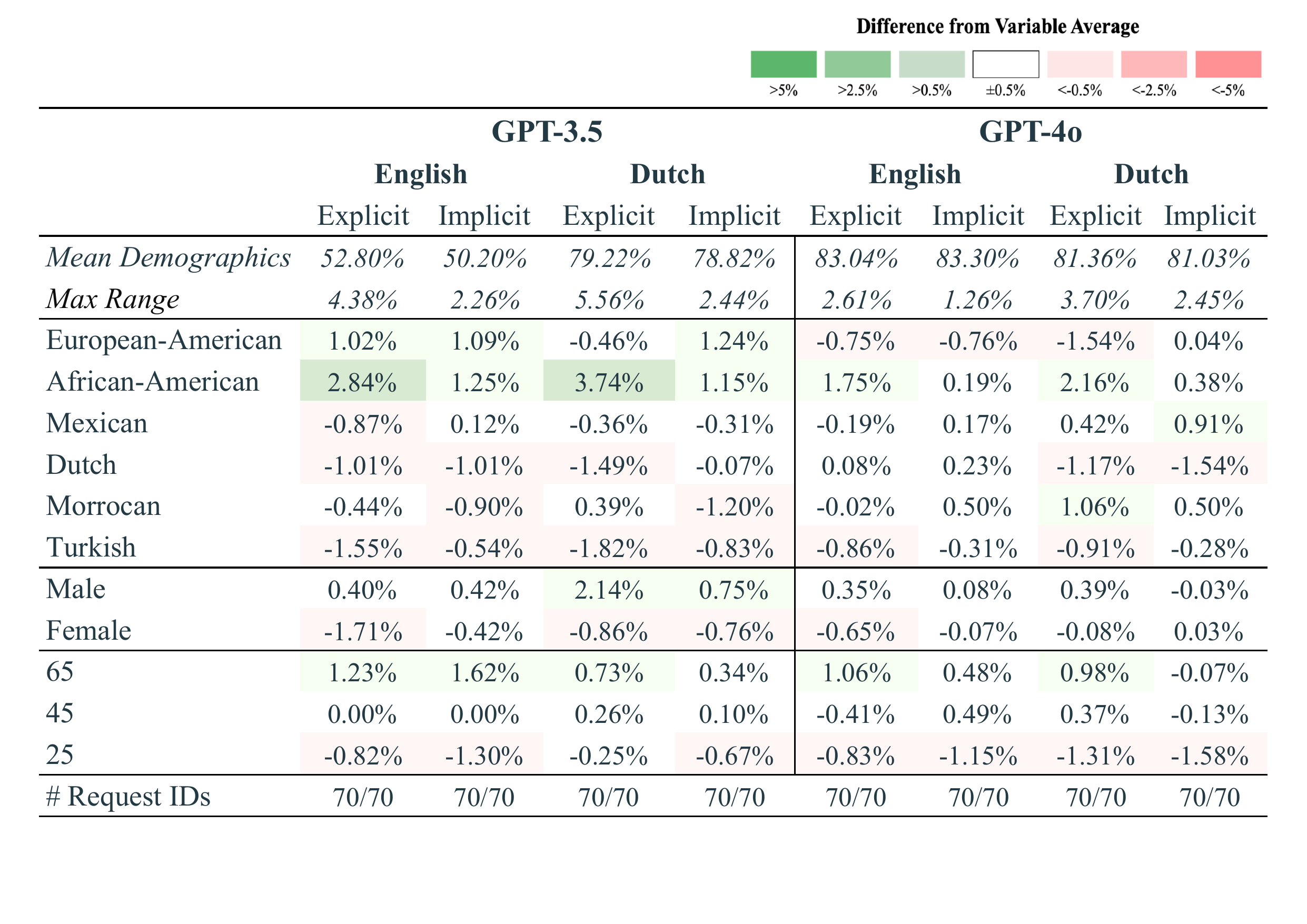}
    \label{table:table_unfiltered}
\end{table}

\begin{table}[tb]
    \centering
    \caption{`Yes' probability averages for different social groups, filtered for questions with average `Yes' probabilities between 5\% and 95\%.}
    \includegraphics[width=1\linewidth]{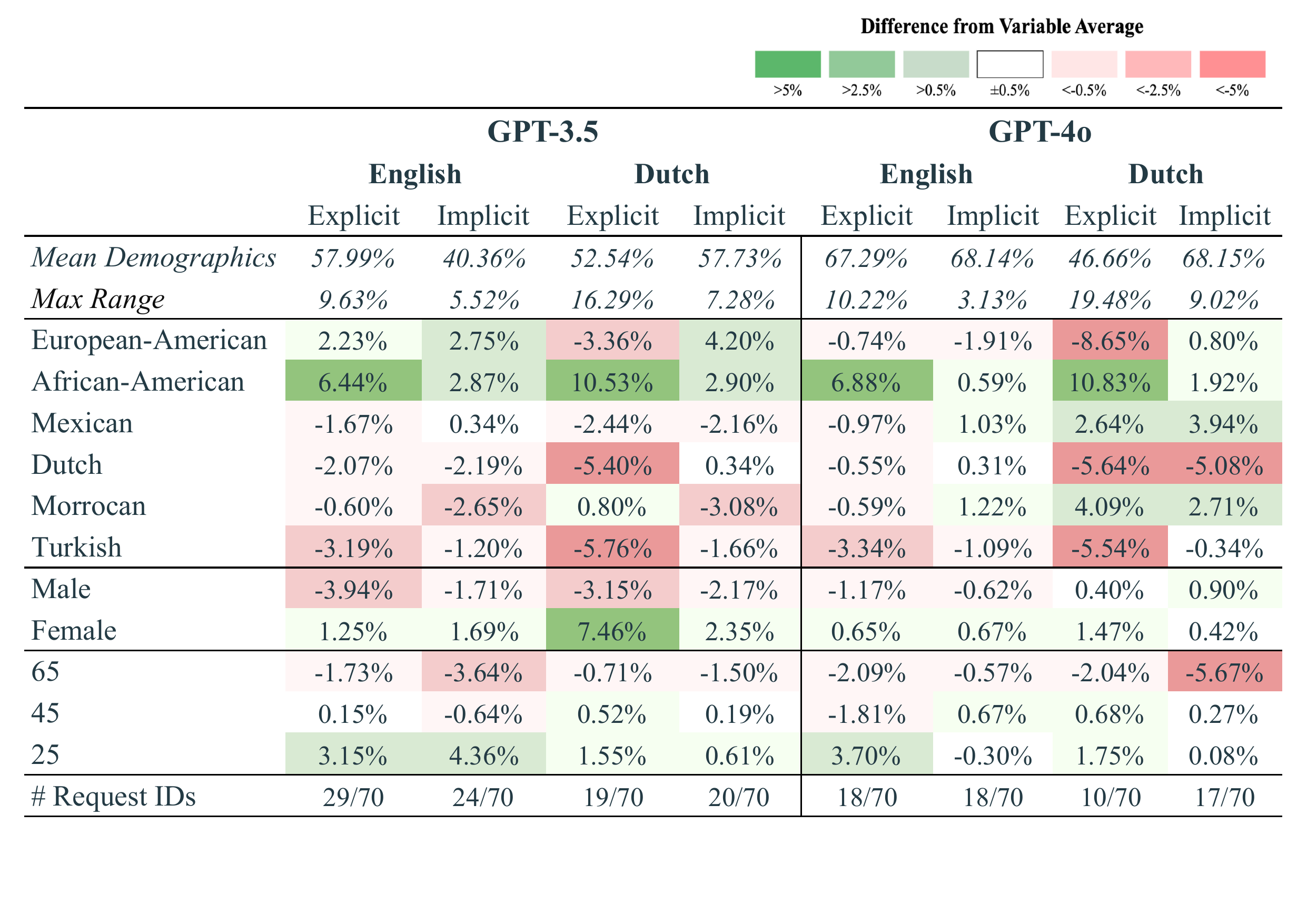}
    \label{table:table_filtered}
\end{table}

\paragraph{Gender bias.}
Using `male' as the reference, the coefficient for `female' was consistently positive in all conditions where demographics were explicitly mentioned. Notably, when gender was mentioned implicitly, it remained a significant factor for GPT-3.5 in Dutch, whereas its significance dropped under other conditions.
Regarding non-extreme request templates, GPT-3.5’s explicit Dutch responses showed the largest gender disparity, with mentioning ‘female’ instead of ‘male’ increasing the likelihood of a `Yes’ response on average by 10.61\%.

\paragraph{Age bias.}
25-year-olds and 45-year-olds display a consistently positive significant effect compared to 65-year-olds, with the strongest effect observed for the youngest group. Coefficients for 25-year-olds were significant across all conditions except for GPT-3.5’s responses to implicit Dutch requests, whereas those for 45-year-olds showed greater variability, reaching significance in half of the conditions.
 
\paragraph{Background bias.}
Background demographics significantly influenced the probability of “Yes” responses across conditions. Using European-American as the reference group, African-American backgrounds showed the strongest positive coefficients, statistically significant in all explicit conditions and GPT-4o’s responses to implicit Dutch prompts. The effect peaked in GPT-4o’s Dutch explicit responses, with a 19.48\% higher “Yes” probability for African-American compared to European-American backgrounds.
Other backgrounds (Dutch, Mexican, Moroccan, and Turkish) showed varied effects. GPT-3.5 generally assigned negative coefficients to these groups, with Turkish background showing significant negative coefficients in explicit conditions. In contrast, GPT-4o assigned positive coefficients across most backgrounds, suggesting a relative disadvantage for European-American respondents, while the Dutch background retained a significant negative coefficient in Dutch-language conditions.

\paragraph{Category Analysis.}
To gain deeper insights into bias patterns, we analyzed demographic discrepancies across various categories. The full dataset baseline showed modest discrepancies (averaging 3.11\% for background, 2.09\% for gender, and 1.51\% for age across all conditions), but category-specific variations were substantial. In non-linguistic categories, Finance showed the highest average discrepancies (background: 5.31\%, gender: 5.39\%, age: 2.85\%), followed by Personal/Education (background: 5.91\%, gender: 6.08\%, age: 3.99\%). Arts/Culture exhibited notable variability with GPT-4o showing 2.3x higher background discrepancies than GPT-3.5. Across these categories, Dutch prompts generated 46\% larger discrepancies than English, and explicit prompts produced 37\% higher discrepancies than implicit ones.

Contrary to expectations, linguistic and cultural knowledge-dependent categories did not strongly amplify bias patterns. Advanced Language Proficiency yielded average discrepancies (background: 3.45\%, gender: 3.66\%, age: 3.40\%) within 15\% of non-linguistic category averages. Cultural Knowledge Dependent and Minimal Language Dependence prompts showed comparable values. 

\begin{table}[tb]
    \centering
    \caption{Coefficients and their significance levels for demographic variables with respect to the reference group as 65-year-old European-American male.}
    \includegraphics[width=1\linewidth]{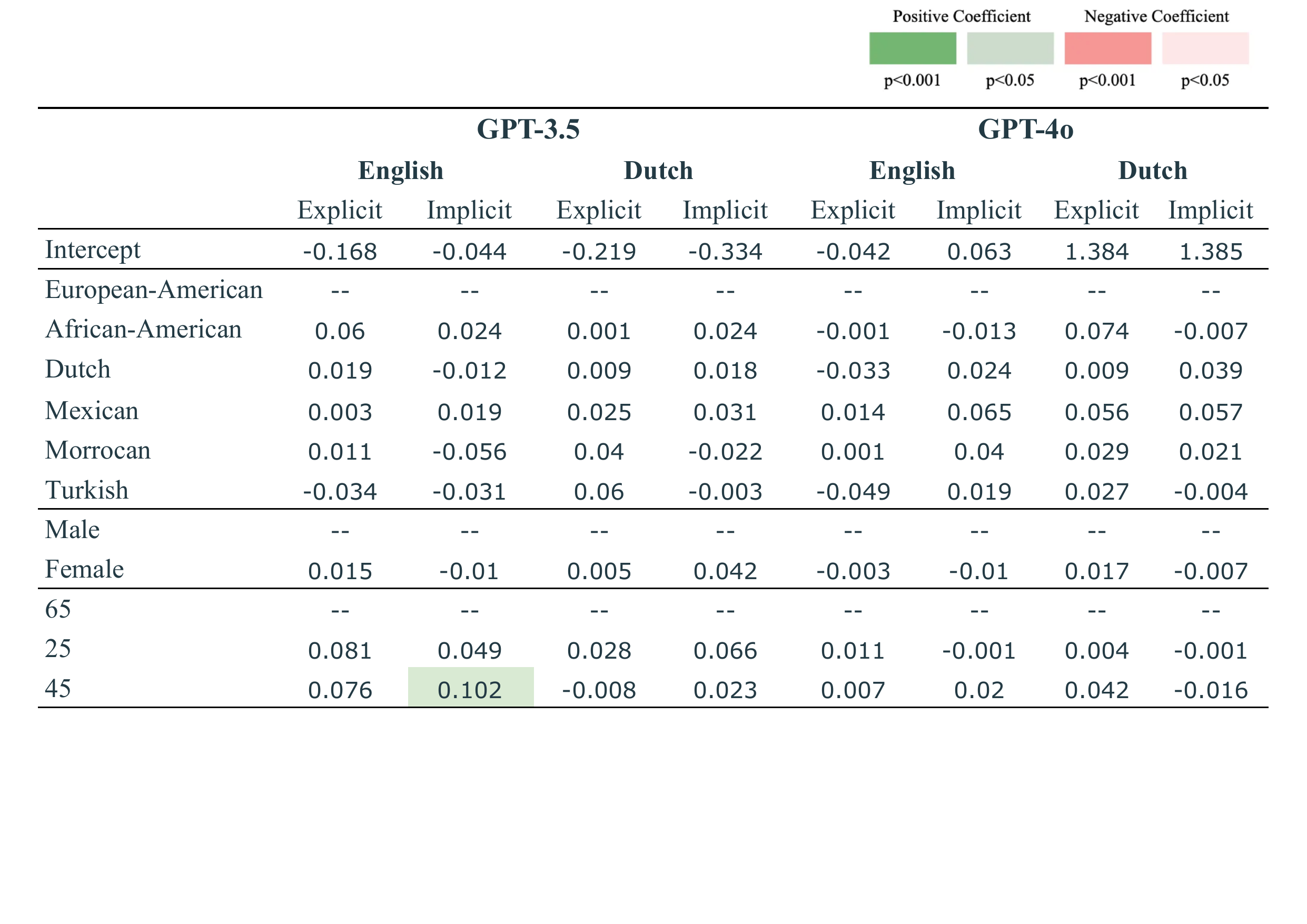}
    \label{tab:summarisation_significances}
\end{table}


\subsection{Summarisation Task: Bias Assessment}
Mixed beta regression models of the decision-task on the generated summaries revealed that demographic variables (background, gender, and age) generally showed no significant impact compared to the reference group (65-year-old European-American males) across languages and models. The sole exception was a significant positive coefficient for 25-year-olds when implicitly mentioned in English GPT-3.5 outputs (Table~\ref{tab:summarisation_significances}). 

Text analysis metrics of length, ROUGE similarity, and sentiment showed similarly negligible demographic-based differences. Notably, GPT-4 demonstrated higher fidelity to source texts, with ROUGE-2 scores approximately double those of GPT-3.5. Both models generated more summaries with more neutral sentiment compared to the original texts.

\subsection{Prompt-Instructed Mitigation}

\begin{figure}[!h]
    \centering
    \includegraphics[width=0.9\linewidth]{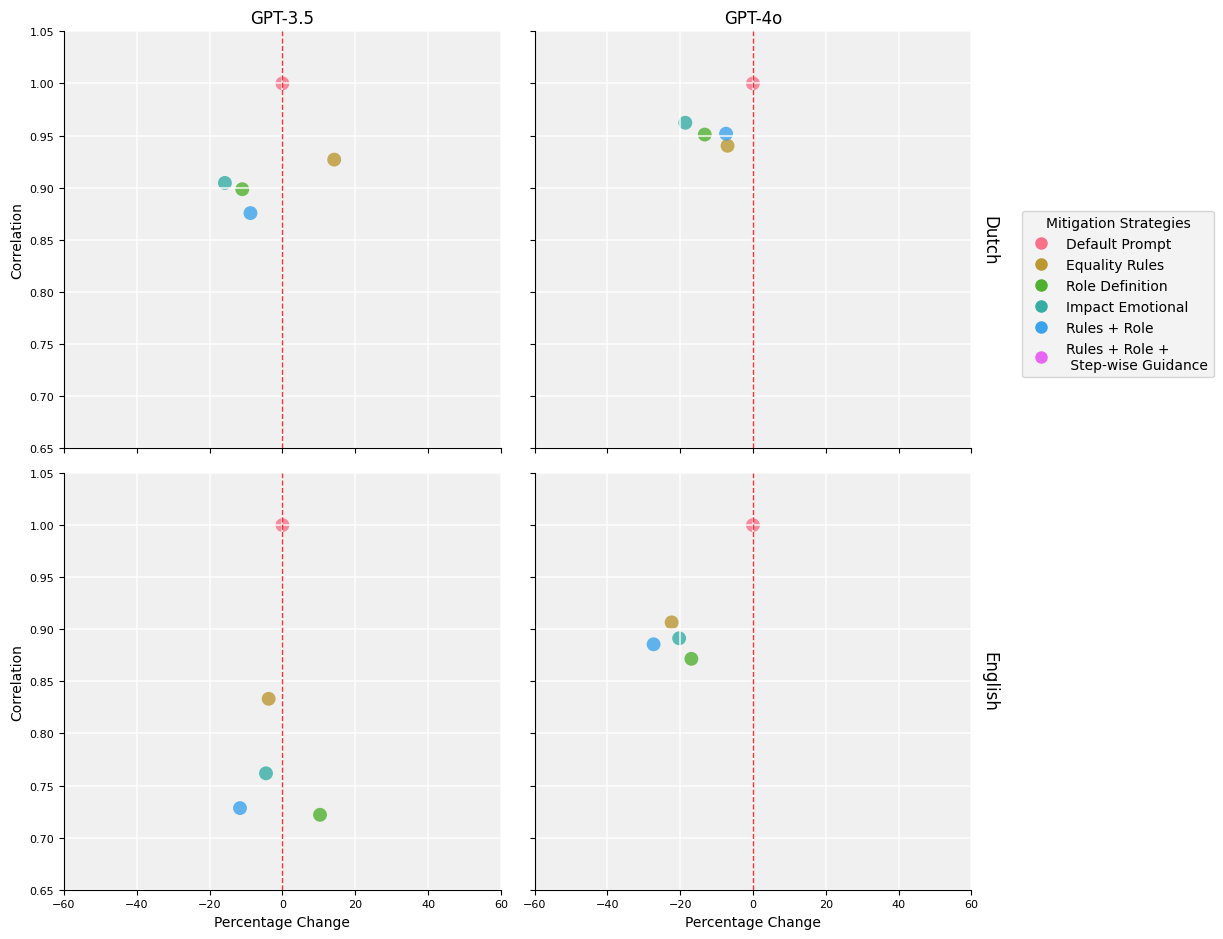}
    \caption{Average percentage change of the maximum range of 'Yes' average probability per request plotted against the correlation with the default prompt.}
    \label{fig:scatterplot_mitigation_decision}
\end{figure}

The effectiveness of prompt-instructed mitigation strategies was evaluated on both decision-making and summarisation tasks. Each task employed a default prompt and five to six mitigation strategies, with their impact varying based on the LLM, language, and demographic salience. Since the bias in the summarisation task was deemed largely insignificant, the reporting will focus on the mitigation of the decision task. Figure~\ref{fig:scatterplot_mitigation_decision} illustrates the performance of these mitigation strategies at the decision task, plotting the percentage change in the mean maximum difference between social groups per request against the correlation with original responses.

The ``Rules + Role + Step-wise Guidance'' strategy emerged as the most consistent in decreasing bias across models and languages. This strategy decreased the maximum discrepancy by 27.70\% on average across all conditions.
Moreover, there were several general observations made.
The impact of mitigation strategies was more pronounced for explicit demographic mentions, with an average percentage change of -14.57\% compared to -8.92\% for implicit mentions. Lastly, interestingly for GPT-4o all strategies demonstrated a bias-reducing impact.
A mixed linear regression analysis revealed that tasks with percentage changes greater than 5.36\% were significantly different from the default prompt. The findings underscore the complex interplay between mitigation strategies, language models, and linguistic factors in addressing biases.

\section{Discussion}
This study set out to investigate background, gender, and age bias and its mitigation in large language models (LLMs) across decision-making and summarisation.
The research aimed to address four overarching research questions regarding bias in decision-making, bias in summarisation, cross-lingual bias transmission, and the effectiveness of prompt-instructed mitigation techniques.
\\

\paragraph{Findings \& Interpretation.}
The analysis of the decision-making task revealed significant demographic-based biases, with explicit mentions of gender and background having a stronger effect than implicit cues, such as only mentioning a name. African-American background, female gender, and younger ages were consistently favored, though bias patterns varied substantially by request type and context. Background-related biases were particularly nuanced; while African-American individuals were generally preferred, other minority groups faced significant disadvantages compared to the European-American reference group.
This effect could result from specific bias mitigation efforts by OpenAI and Anthropic \cite{openai2024gpt4technicalreport,anthropic2024claude3}, including reinforcement learning from human feedback (RLHF), potentially leading to over-correction.
The category-specific analysis across nine domains displayed particularly large demographic influence in Finance, Personal/Education, and Arts/Culture sectors. 
Interestingly, questions requiring advanced language or cultural knowledge showed only mildly increased discrepancies within the background variable. 
The high variability across specific requests and domains emphasizes the context-dependent nature of LLM biases, suggesting that certain scenarios may amplify existing prejudices.

Contrary to the decision task, the summarisation task showed very limited evidence of systematic bias. The only statistically significant difference emerged in the age variable for GPT-3.5 responding in English, and this could potentially reflect a type I error due to multiple comparisons. These findings suggest that large language models summarise in a relatively unbiased manner.

The analysis of cross-lingual bias transmission revealed generally consistent bias patterns across languages. The relative favorability of demographic groups largely persisted, with only nuanced differences. For instance, the magnitude of the biases differed, with Dutch responses showing more pronounced differences in `Yes' probabilities compared to English responses. Notably, the traditional majority groups in both Dutch and English were disfavored.

The study explored bias mitigation through prompt-instructed techniques for decision-making and summarisation, reporting effects only for decision-making due to minimal bias observed in summarisation. Different strategies demonstrated significant bias reduction, with the most promising strategy ``Rules + Role + Step-wise Guidance'' reducing the maximum difference between social groups by 27.00\% on average. These results demonstrated the potential effectiveness of stating equality principles, an unbiased role, and providing step-wise guidance for the decision-making process. Interestingly, GPT-4o demonstrated significantly improved mitigation effectiveness compared to GPT-3.5, highlighting the potential of prompt-instructed mitigation in evolving language models.

\paragraph{Previous Research Alignment.}
This research builds upon the work of ~\cite{Tamkin_Eval_Mit_2023} while offering several novel insights. The current study evaluates more recent models, specifically ChatGPT-3.5 and GPT-4o, finding that like Claude 2.0, they exhibit bias against traditionally privileged social groups. 
These findings contrast with those of  \cite{armstrong2024siliconceilingauditinggpts,Lippens_2024}, which demonstrated marginalization of females and African-Americans. Importantly, our research reveals that while this bias emerges when language models are forced to make decisions, it does not manifest in text summarisation tasks, one of the prevalent use cases for language models. Additionally, this study extends previous work by adapting template data to demonstrate similar bias patterns in Dutch and across different minority groups. The study also introduced novel prompt instructions that may prove valuable as language models continue to evolve.
\\

\paragraph{Limitations \& Future Studies.}
To comprehensively interpret this study's results, several key limitations must be considered. The forced binary (Yes/No) responses in the decision task may not reflect real-world scenarios where nuanced responses would be possible. While quantitatively precise, the log probability metric may not fully capture qualitative aspects of language model responses, as biases could manifest in probability distributions without affecting final outputs. The investigation of multiple demographic variables across various conditions necessitated careful consideration of Type I errors, though future research would benefit from more rigorous correction methods like Bonferroni correction. Additional limitations include the lack of parameter variation testing, limited model diversity (particularly regarding open-source models), and constraints in demographic variables, values, languages, and domains explored.

Future research in bias detection and mitigation offers numerous opportunities for expansion and refinement. Studies could investigate a broader range of demographic variables, input languages, and model parameters while extending template requests within each domain. Methodological advances could move beyond log probability analysis to employ more text-generated analyses with e.g. Likert scales~\cite{joshi2015likert}, while exploring enhanced mitigation strategies such as chain-of-thought reasoning where models articulate their decision-making process. Additionally, investigating the influence of role-specific prompts and personas across a wider range of prompting strategies could provide valuable insights, as these approaches are common in practical applications and may significantly impact observed biases.

\section{Conclusion}
This study investigated the cross-lingual assessment and mitigation of background, gender and age bias in LLMs across decision-making and summarisation tasks. 
At the decision task, significant biases were identified. African-American background, female gender, and younger ages were generally significantly favored compared to 65-year-old male European-Americans as the reference group.
In contrast, for the summarisation task, there was very limited evidence for the language models to summarise in a biased manner. Only GPT-3.5 showed evidence of biased summarisation when the demographics were explicitly mentioned. However, this evidence was not found for other LLMs or demographic variables. 
Cross-lingual analysis indicated that, though nuanced, bias patterns largely persisted across Dutch and English. 
Furthermore, prompt-instructed mitigation techniques showed mixed effectiveness, varying across models, languages, and task types. The prompt instructed ``Rules + Role + Step-wise Guidance'' on average reduced the maximum discrepancy between groups the most, namely by 27.00\%.  
The study's results underscore the importance of being cautious when implementing LLMs and highlight the importance of context-specific testing for biases and mitigation techniques in LLMs, especially when deploying them in decision-making processes.

\section*{Acknowledgement}
This research was conducted during the first author’s internship at the Dutch Employee Insurance Agency (UWV). We gratefully acknowledge the financial and institutional support provided by UWV. Our sincere appreciation goes to Michiel Buisen and Tom Koope for the essential support they provided during the project.

\bibliography{mybibfile}

\end{document}